\documentclass[conference]{IEEEtran}
\ifCLASSINFOpdf
\else
\fi
\usepackage{amsmath, algorithm, algorithmicx, algpseudocode, graphicx, bbm}
\begin{document}
\title{Reducing the Model Order of Deep Neural Networks Using Information Theory}

\author{\IEEEauthorblockN{Ming Tu$^{1}$, Visar Berisha$^{1, 2}$, Yu Cao$^2$, Jae-sun Seo$^{2}$,}
\IEEEauthorblockA{$^{1}$ Speech and Hearing Science Department, Arizona State University \\
    $^{2}$ School of Electrical, Computer, and Energy Engineering, Arizona State University}}


%Telephone: (800) 555--1212, Fax: (888) 555--1212}


\maketitle

\begin{abstract}
Deep neural networks are typically represented by a much larger number of parameters than shallow models, making them prohibitive for small footprint devices. Recent research shows that there is considerable redundancy in the parameter space of deep neural networks. In this paper, we propose a method to compress deep neural networks by using the Fisher Information metric, which we estimate through a stochastic optimization method that keeps track of second-order information in the network. We first remove unimportant parameters and then use non-uniform fixed point quantization to assign more bits to parameters with higher Fisher Information estimates. We evaluate our method on a classification task with a convolutional neural network trained on the MNIST data set. Experimental results show that our method outperforms existing methods for both network pruning and quantization.
\end{abstract}


% \fi
\IEEEpeerreviewmaketitle

\section{Introduction}

Deep neural networks (DNNs) have been shown to outperform shallow learning algorithms in applications such as computer vision \cite{Simonyan14c, szegedy2015going}, automatic speech recognition \cite{hinton2012deep, graves2014towards} and natural language processing \cite{lecun2015deep, goldberg2015primer}; however DNNs also have large parameter sets, often making them prohibitive for small-footprint devices \cite{gokhale2014240}. For example, while the original LeNet5 network \cite{lecun1998gradient} (a classification system based on convolutional neural network) has less than 100K parameters, the winner of the 2012 ImageNet competition \cite{krizhevsky2012imagenet} has over 60M parameters. The memory access costs alone can make these larger networks unsuitable for low-power settings.

It has been posited that the expressive power of DNNs comes from their large parameter spaces and hierarchical structure; however recent studies have shown that there is often a great deal of parameter redundancy in DNNs \cite{cheng2015exploration, denil2013predicting}, making them unnecessarily complex. As a result, reducing the complexity of DNNs has been an area of great interest to the research community in recent years. For example, the authors in  \cite{xue2013restructuring, denton2014exploiting} used low rank decomposition of the weights to reduce the parameter set and applied this method to a DNN-based acoustic model and to convolutional neural networks (CNN) for image classification. Similarly, the authors in \cite{denil2013predicting} showed that over 95\% parameters of DNNs can be predicted without any training and without impacting accuracy.

In addition to low-rank parameter decomposition, network pruning and quantization methods have also been proposed \cite{han2015learning}. Neural network pruning has been investigated in early studies, including pruning weights with small magnitudes, optimal brain damage \cite{lecun1990optimal} and optimal brain surgeon \cite{hassibi1993second}. The last two methods require estimation of the Hessian matrix of network parameters to decide on their importance; however, the sizes of existing networks make the estimation of this large matrix prohibitive. As a result, for large-scale DNNs, magnitude-based weight pruning is still a popular method \cite{yu2012exploiting, han2015learning, han2015deep}.

For fixed-point implementations of DNNs, parameter quantization is also required. The studies in \cite{koksal2001weight, takeda2014boundary} discretized the weights of a neural network according to the range of the weights. The methods in \cite{lei2013accurate} and \cite{vanhoucke2011improving} used uniform scalar parameter quantization to implement fixed-point versions of the networks. In \cite{gupta2015deep}, a new fixed-point representation for DNN training was proposed, using stochastic rounding for the parameters. Vector quantization based schemes have been applied to CNNs for both computer vision and automatic speech recognition tasks \cite{gong2014compressing, soulie2015compression, wang2015small}.

In this paper, we propose a new method that ranks the parameters of a DNN for both network pruning and parameter quantization. We investigate an information-theoretic approach to reduce the DNN parameter space by using the \textit{Fisher Information} as a proxy for parameter importance. The Fisher criterion is a natural metric for quantifying the relative importance of DNN parameters since it provides an estimate of how much information a random variable carries about a parameter of the distribution. In \cite{tu2016ranking}, we introduced a new method to calculate the diagonal of the Fisher Information Matrix (FIM) and showed that it can be used to reduce the size of DNNs. In this paper, we extend this work by using a lower-complexity estimate of the FIM diagonal and evaluating the technique on a much larger network. We validate the method on the MNIST dataset using a CNN and show that our method results in smaller networks with fewer parameters to quantize at a lower bit rate.

 The remainder of this paper is organized as follows: In the next section, we introduce our network pruning and quantization scheme. In section \ref{exp}, we validate the algorithm on the MNIST data. We end the paper with a discussion of the results in section \ref{discussion} and concluding remarks in section \ref{conclusion}.

%  and prune neural network parameters with low fisher information. With experimental results on an autoencoder, it has been proved Fisher Information can provide better pruning results than magnitude based pruning, especially for parameters with relatively large magnitude. One drawback of this method is that though highly parallizable this method could still be slow for large-scale DNNs. Here, we use another method to approximate diagonals of FIM. Our method is based on gradient descent based optimization methods for DNNs which utilize the geometry information of the network. Thus, Fisher Information can be estimated during training without any after-training computation, making this method as accessible as magnitude based pruning but can provide more compression. Also, for parameters survive pruning, we provide another method to do \textit{non-uniform} quantization. Parameters with high Fisher Information will be assigned more bits and vice versa.

\begin{figure}[t]
\minipage{0.46\textwidth}
  \includegraphics[width=\linewidth]{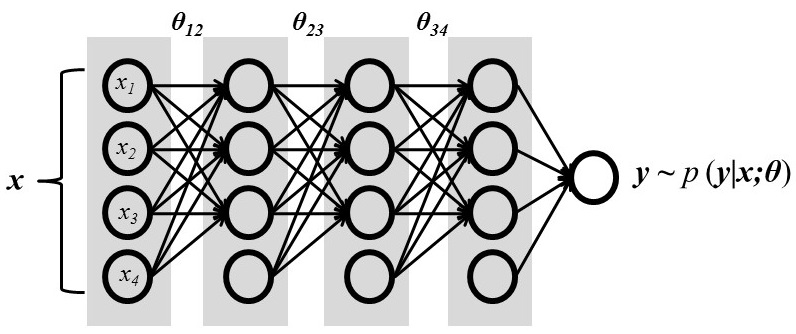}
  \caption{A notional DNN architecture with input $\mathbf{x}$, output $\mathbf{y}$, and parameter vector $\boldsymbol{\theta}$.}\label{dnn}
\endminipage
\end{figure}

\section{DNNs pruning and quantization}
\label{method}

\subsection{Fisher Information and DNNs}
We show a notional DNN architecture in Fig. \ref{dnn}. Let us consider the output $\mathbf{y}$ of the DNN as a conditional probability distribution $p(\mathbf{y}|\mathbf{x};\boldsymbol\theta)$ parameterized by the DNN input, $\mathbf{x}$, and its parameters, $\boldsymbol\theta$. The FIM evaluated at a particular value of $\boldsymbol\theta$ is defined as:

\begin{equation}
\label{FIMeqn}
\mathbf{F}({\boldsymbol{\theta}}) = \mathrm{E}_{\mathbf{y}} \left [ \left ( \frac{\partial \ \mathrm{log} \ p(\mathbf{y}|\mathbf{x};\boldsymbol\theta) }{\partial \ \boldsymbol\theta} \right ) \left ( \frac{\partial \ \mathrm{log} \ p(\mathbf{y}|\mathbf{x};\boldsymbol\theta) }{\partial \ \boldsymbol\theta} \right )^T \right ].
\end{equation}

We can see from eqn. (\ref{FIMeqn}) that the FIM is the covariance of the gradient of log likelihood with regards to its parameter $\boldsymbol\theta$. Thus, $\mathbf{F}_{\boldsymbol{\theta}}$ is a $p \times p$ symmetric positive semidefinite matrix.

It is easy to see that the diagonal elements of the FIM can be calculated by the expectation of the element-wise multiplication of the gradient:
\begin{equation}
\mathbf{F}_{D}({\boldsymbol{\theta}}) = \mathrm{E}_{\mathbf{y}} \left [ \mathbf{g} \odot \mathbf{g} \right ],
\end{equation}
where $\mathbf{g}=\frac{\partial \ \mathrm{log} \ p(\mathbf{y}|\mathbf{x};\boldsymbol\theta) }{\partial \ \boldsymbol\theta}$ is the gradient of the log-likelihood and $\odot$ represents element-wise multiplication; $\mathbf{F}_{D}(\boldsymbol{\theta})$ is a $p \times 1$ vector, each element of which is the Fisher Information of a corresponding parameter.

The Fisher Information provides an estimate of the amount of information a random variable carries about a parameter of the distribution. In the context of a DNN, this provides a natural metric for quantifying the relative importance of any given parameter in the network. The less information an output variable carries about a parameter, the less important that parameter is to the output statistics of the network. As a result, we assume that removing parameters with low entries in $\mathbf{F}_{D}({\boldsymbol{\theta}})$ will not greatly affect the output of the network. That is precisely the approach we take in this paper - we will rank the parameters in a DNN based on their corresponding entries in the FIM diagonal. In the ensuing section, we describe the method we use for approximating $\mathbf{F}_{D}({\boldsymbol{\theta}})$.

\subsection{Estimating the Fisher Information}

A number of recent studies on natural gradient descent (NGD) \cite{amari1998natural} exploit the information geometry of the underlying parameter manifold and apply it to gradient-based optimization of DNNs. Natural gradient descent uses the inverse FIM to constrain the magnitude of the update steps such that the Kullback-Leibler (KL) divergence between the output distribution of the network at iteration $t$ and iteration $t+1$ is constant \cite{pascanu2013revisiting, povey2014parallel, desjardins2015natural}. This approach avoids large update steps and results in faster convergence.

For classification problems, DNNs are trained by minimizing the cross-entropy loss function:
\begin{equation} \label{eqn:crossentr}
f(\boldsymbol\theta)=-\sum_{i=1}^{N}\sum_{k=1}^{C}\mathbbm{1}(y^{(i)}=k)log(p(y^{(i)}|\mathbf{x}^{(i)};\boldsymbol\theta)),
\end{equation}
where $N$ is the number of training samples, $C$ is the number of classes, $y^{(i)}$ is the true label of $i^{\mathrm{th}}$ sample $\mathbf{x}^{(i)}$ and $\mathbbm{1}\left \{ \cdot  \right \}$ is the indicator function. A recent paper proposed a new stochastic optimization method called ``Adam'' and showed we can efficiently estimate the FIM diagonal at each iteration while minimizing this loss function \cite{kingma2014adam}. Similar to NGD, Adam uses the approximated FIM diagonal to adapt to the geometry of the data. As a result, in this study, we use Adam to train our DNN classification system. The details of the parameter update scheme for Adam are shown in algorithm \ref{alg:adam}. As the algorithm shows, after the training algorithm converges, it returns both the optimal $\boldsymbol\theta_t $ and the approximated Fisher Information $\mathbf{\hat{F}}_{D}({\boldsymbol{\theta}_t})$. We should note that Adam is not the only choice as the optimizer because standard stochastic gradient descent can also be used; however this would require some other means of estimating $\mathbf{\hat{F}}_{D}({\boldsymbol{\theta}_t})$.

\begin{algorithm}[t]
\caption{Adam algorithm, excerpted from \cite{kingma2014adam}}
\label{alg:adam}
\begin{algorithmic}
\State \bf{Require:} step size $\alpha$, exponential decay rates $\beta_1$, $\beta_2$, $\epsilon$
\State Given initial parameter vector $\boldsymbol\theta_0$, initial first and second moment vectors $\boldsymbol m_0 \leftarrow 0$ and $\boldsymbol v_0 \leftarrow 0$ and initial timestep $t$
\State While $\boldsymbol \theta_t$ not converged do:
\State \,\,\,\,\,\,\,\, 1. $t \leftarrow t+1$
\State \,\,\,\,\,\,\,\, 2. $\boldsymbol g_t \leftarrow \nabla_{\boldsymbol \theta} f_t(\boldsymbol \theta_{t-1})$ (Get gradients)
\State \,\,\,\,\,\,\,\, 3. $\boldsymbol m_t \leftarrow \beta_1 \boldsymbol m_{t-1}+(1-\beta_1) \boldsymbol g_t$
\State \,\,\,\,\,\,\,\, 4. $\boldsymbol v_t \leftarrow \beta_2 \boldsymbol v_{t-1}+(1-\beta_2) \boldsymbol g_t \odot \boldsymbol g_t$
\State \,\,\,\,\,\,\,\, 5. $\boldsymbol{\hat{m}_t} \leftarrow \boldsymbol m_t / (1-\beta_1^t)$
\State \,\,\,\,\,\,\,\, 6. $\mathbf{\hat{F}}_{D}({\boldsymbol{\theta}_t}) \leftarrow \boldsymbol v_t / (1-\beta_2^t)$
\State \,\,\,\,\,\,\,\, 7. $\boldsymbol \theta_t \leftarrow \boldsymbol \theta_{t-1}-\alpha \boldsymbol{\hat{m}_t}/(\mathbf{\hat{F}}_{D}({\boldsymbol{\theta}_t})+\epsilon)$
\State end while
\State return $\boldsymbol\theta_t $, $\mathbf{\hat{F}}_{D}({\boldsymbol{\theta}_t})$

\end{algorithmic}
\end{algorithm}

%We use Adam to train our DNN and to estimate the Fisher information. $\mathbf{\hat{F}}_{D}({\boldsymbol{\theta}_t})$ in algorithm \ref{alg:adam} has been proved to be the approximated FIM diagonals \cite{kingma2014adam}.

%We use the Keras \cite{chollet2015keras} implementation of Adam and apply it to CNN based handwriting digits classification on MNIST dataset. We should note that Adam is not the only choice as optimizer because we can even use standard stochastic gradient descent to realize our method just by adding the tracking of $\mathbf{\hat{F}}_{D}({\boldsymbol{\theta}_t})$.

\subsection{Network Pruning and Quantization} \label{sec:pruquan}

The simplest approach to network pruning is to rank the parameters by comparing their entries in the FIM diagonal and removing the ones with the lowest entries. However, as we will see in the results section, this method does not work well since estimating small values in the FIM diagonal is challenging and unreliable. When the model is over-parameterized, the actual parameter space is much smaller than the number of parameters used in the network. As a result, after training, a number of parameters become close to zero and estimating their influence based on the Fisher Information is challenging \cite{tu2016ranking}. To address this problem we use a combination of magnitude-based and FIM-based pruning. For example, if we want to prune $L$ parameters from the network, we first remove $L(1-r)$ parameters with the smallest magnitude. Then, we rank the remaining parameters based on their FIM diagonal entries and remove the additional $Lr$ parameters with the smallest entries in the FIM diagonal. The parameter $r$ is between 0 and 1 and can be optimized using cross-validation.

After network pruning, we want to quantize the remaining parameters with the lowest bit representation possible. After removing $L$ parameters, we rank the remaining $p-L$ parameters by comparing their entries in the FIM diagonal and then apply $k$-means clustering to separate the parameters into several groups. We quantize groups with higher Fisher Information values using more bits and groups with lower Fisher Information using fewer bits.

\section{Experiments and Results Analysis}
\label{exp}

In this section, we present the experiments and results in two parts: network pruning and network quantization. All the experiments were done using the Python neural network library Keras \cite{chollet2015keras} implemented using Theano on an NVIDIA GTX 760 GPU.

We evaluated our algorithms on the MNIST digits data set, which consists of 60K binary images for training and 10K for testing. There are 10 classes (digits from 0 to 9) in the data and the size of each image (and the input dimension of the neural network) is $28 \times 28$. We trained a convolutional neural network (CNN) with 2 convolutional layers each with 32 filters. The size of the convolutional kernel was $3 \times 3$ and the rectified linear unit (ReLU) activation function was used. There was a $2 \times 2$ max-pooling layer following the two convolutional layers with 0.25 dropout probability. Before the output layer, there was a fully connected layer with 128 nodes with ReLU activations and 0.5 dropout probability. The output layer had 10 nodes with softmax activations.

The loss function used to train the network was the categorical cross-entropy shown in eqn. (\ref{eqn:crossentr}). We used Adam as the optimizer with the following settings \ref{alg:adam}: batch size = 256, number of epochs = 50, step size $\alpha$ = 0.001, $\beta_1$ = 0.9, $\beta_2$ = 0.999, $\epsilon=1\mathrm{E}-8$. The accuracy of this model on the MNIST classification task without any pruning and quantization was 99.29\%.

Below we describe the performance of the proposed algorithm for both pruning and quantization tasks. While we focus on CNNs in this section, our method is in no way restricted to only CNNs or only classification networks. Indeed, our probabilistic interpretation of the DNN output makes the methodology applicable across all network types, provided that the Fisher Information can be accurately estimated. Since the fully connected layers of CNNs accounts for $\sim$ 90\% of the total weights \cite{zeiler2014visualizing},  we only focus on the  weights (including bias terms) in the fully connected layers in this paper as others have done in \cite{gong2014compressing} \cite{soulie2015compression}.

\subsection{Network pruning}

\begin{figure}[t]
\minipage{0.46\textwidth}
  \includegraphics[width=\linewidth]{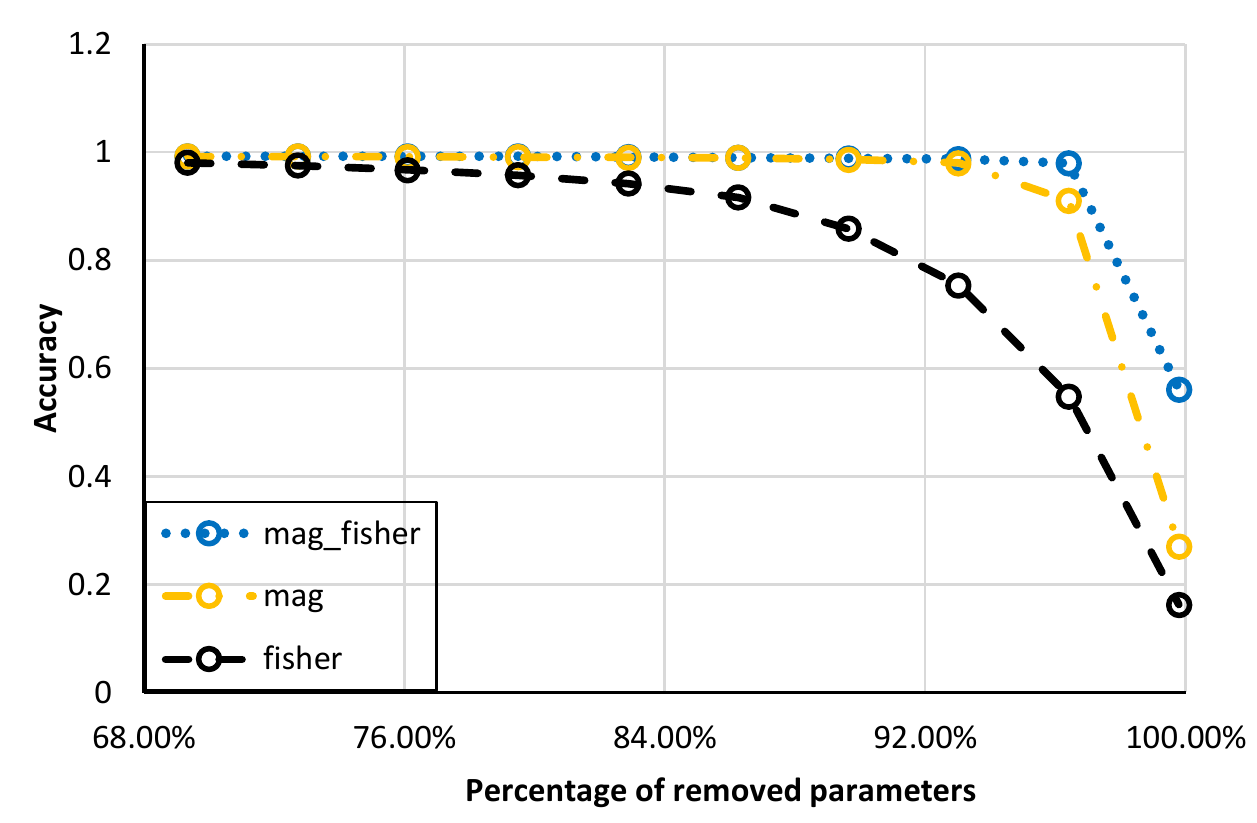}
  \caption{Accuracy change with increasing percentage of removed parameter.}\label{fig1}
\endminipage\newline
\minipage{0.46\textwidth}
  \includegraphics[width=\linewidth]{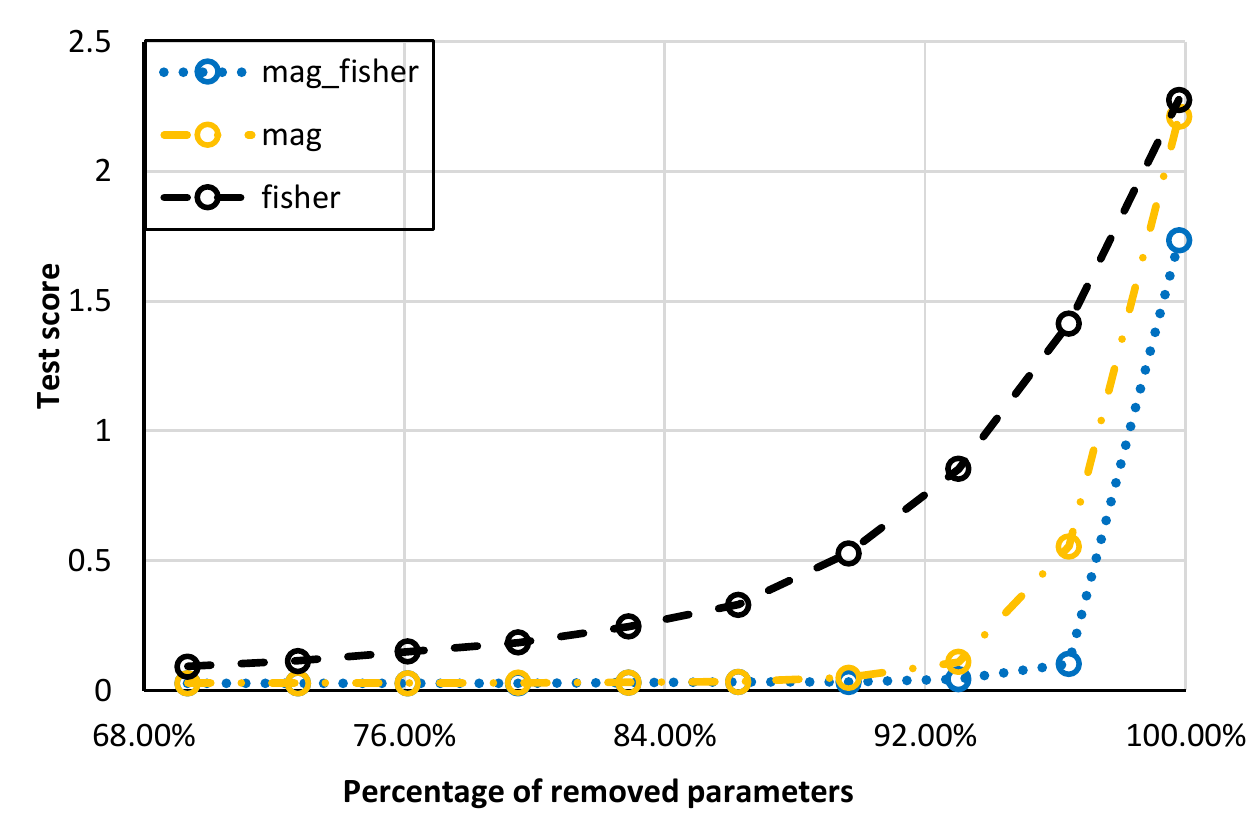}
  \caption{Test score change with increasing percentage of removed parameter.}\label{fig2}
\endminipage\hfill
\end{figure}

The trained network consisted of a total of 591,242 parameters in the fully connected layers. We removed different numbers of parameters using three different methods: (1) magnitude-based pruning where parameters with the smallest magnitude were removed; (2) Fisher Information based pruning where parameters with small entries in the FIM diagonal were removed; and (3) a combination of magnitude and Fisher Information based pruning, where we traded off between the two methods using the parameter $r$ (see Sec. \ref{sec:pruquan}). The number of pruned parameters ranged from $1.0\mathrm{E}4$ (1.69\% of the total parameters) to $5.9\mathrm{E}5$ (99.79\% of the total parameters) with a step size $2.0\mathrm{E}4$. For the third method, we fixed the $r$ value to 0.05. On this network we found that $r$ values below 0.1 yield good results; however for other networks cross-validation could be used to identify an appropriate value of $r$.

\begin{figure}[t]
\minipage{0.46\textwidth}
  \includegraphics[width=\linewidth]{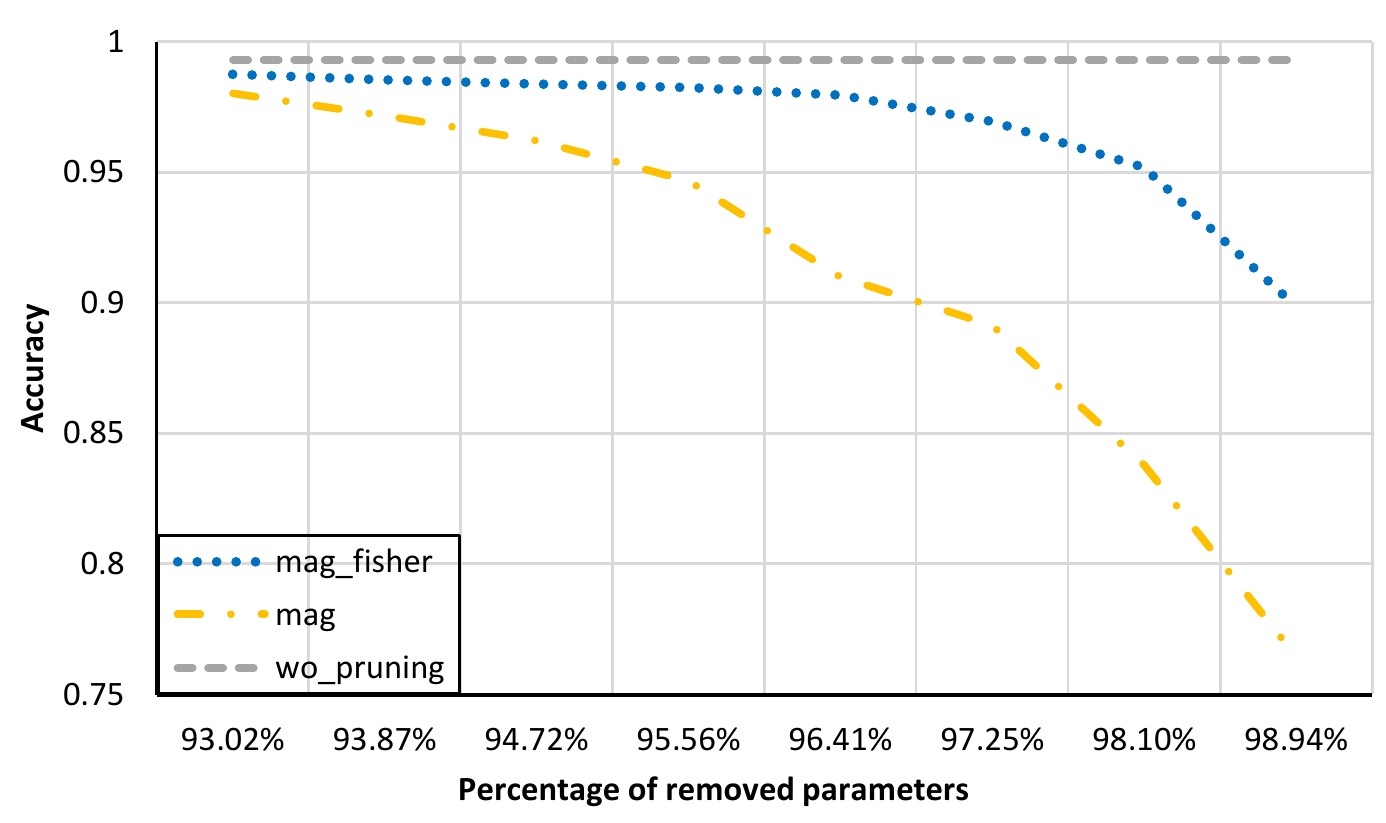}
  \caption{Comparison between proposed pruning method and magnitude based pruning with finer scale.}\label{fig3}
\endminipage
\end{figure}

After removing unimportant parameters in the network, we evaluated the results on the test data and noted both the accuracy of the model and test scores (loss function evaluated on test data) as different numbers of parameters were removed. Since there was no obvious accuracy drop until $4.1\mathrm{E}5$ parameters were removed  (69.35\% of the total parameters), we only show the accuracy and test score starting with 69.35\% parameters removed. The results are shown in Fig. \ref{fig1} and \ref{fig2}. In Fig. \ref{fig1}, we show the accuracy of the model on the test data as we remove an increased number of parameters; In Fig. \ref{fig2}, we show the same plot, but with the test score instead of the accuracy. In these figures, ``mag'' represents magnitude based pruning, ``fisher'' represents Fisher Information based pruning and ``mag\_fisher'' represents a combination of magnitude and Fisher Information based pruning.

 As Fig. \ref{fig1} shows, as additional parameters are removed, the accuracy of the model eventually decreases. We find that using only Fisher Information based pruning, the accuracy of the model decreases quickly. This is because estimating the FIM for small parameter values is difficult as explained in Sec. \ref{sec:pruquan}. This is consistent with our finding in \cite{tu2016ranking}, where we used the FIM criterion to remove parameters in an autoencoder. Using magnitude-based pruning, there is a clear drop-off in performance after $5.5\mathrm{E}5$ parameters (93.02\% of the total parameters) are removed; however, by using our combination of magnitude and Fisher Information pruning there is no obvious accuracy drop until $5.7\mathrm{E}5$ parameters (96.41\% of the total parameters) are removed. The advantage of the combined method shows that about $2.0\mathrm{E}4$ more parameters (3.38\% of the total parameters) can be removed compared to magnitude based pruning with minimal impact on model performance. The test score in Fig. \ref{fig2}  follows the same trend as the accuracy plot in Fig. \ref{fig1}.

To further highlight the differences in performance between ``mag'' and ``mag\_fisher'', we zoom in at the point where the accuracy starts to decrease by running the experiment with a smaller step size. These findings are shown in Fig. \ref{fig3}, where we show the accuracy (99.29\%) without any pruning. The starting point of ``mag\_fisher'' is 98.75\% while for ``mag'' it is 98.02\%. From this result, we can more clearly see the advantage of our combined method compared to magnitude based pruning. Consistent with findings from our previous work \cite{tu2016ranking}, the Fisher Information better captures the importance of larger parameters when compared to magnitude-based pruning.

\subsection{Network quantization}

\begin{figure}[t]
\minipage{0.46\textwidth}
  \includegraphics[width=\linewidth]{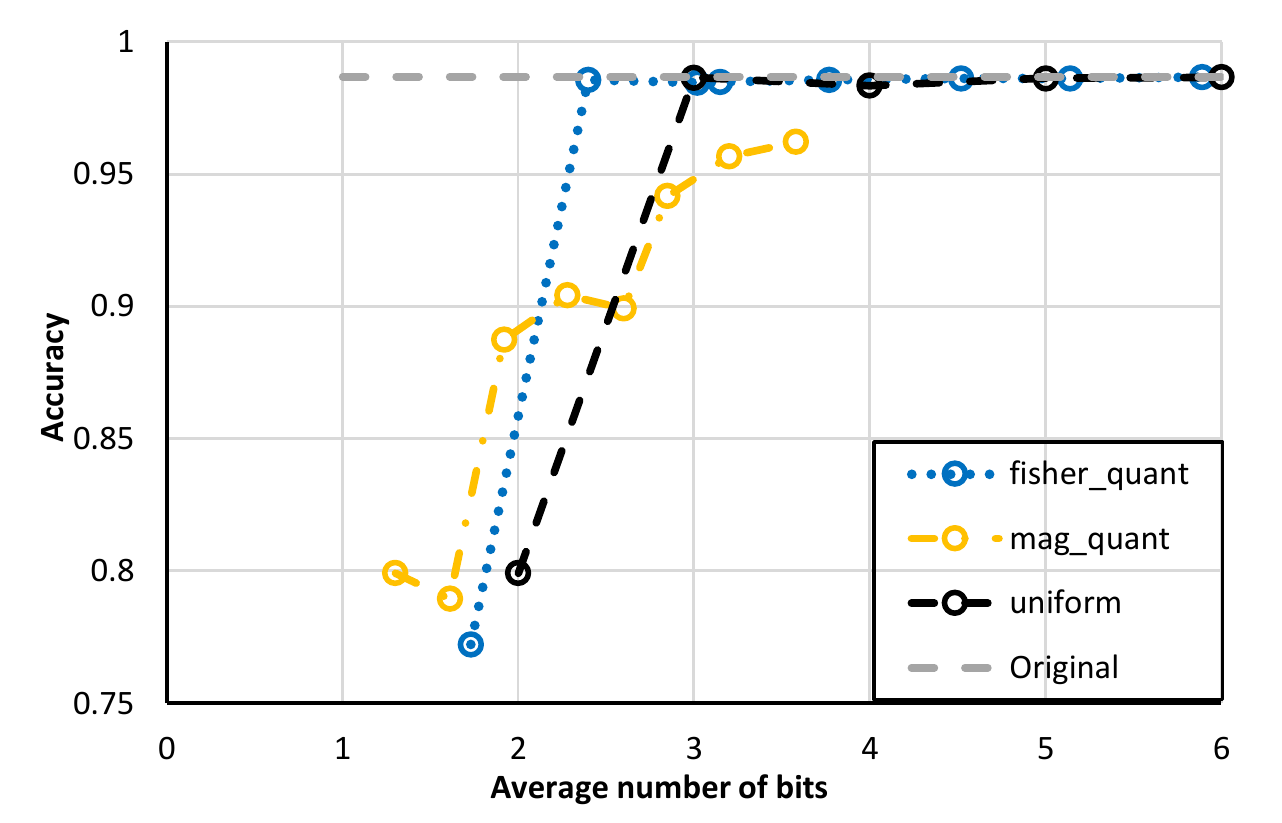}
  \caption{Scatter plot with line to show pairs of accuracy and average number of bits of different quantization methods.}\label{fig4}
\endminipage\newline
\minipage{0.46\textwidth}
  \includegraphics[width=\linewidth]{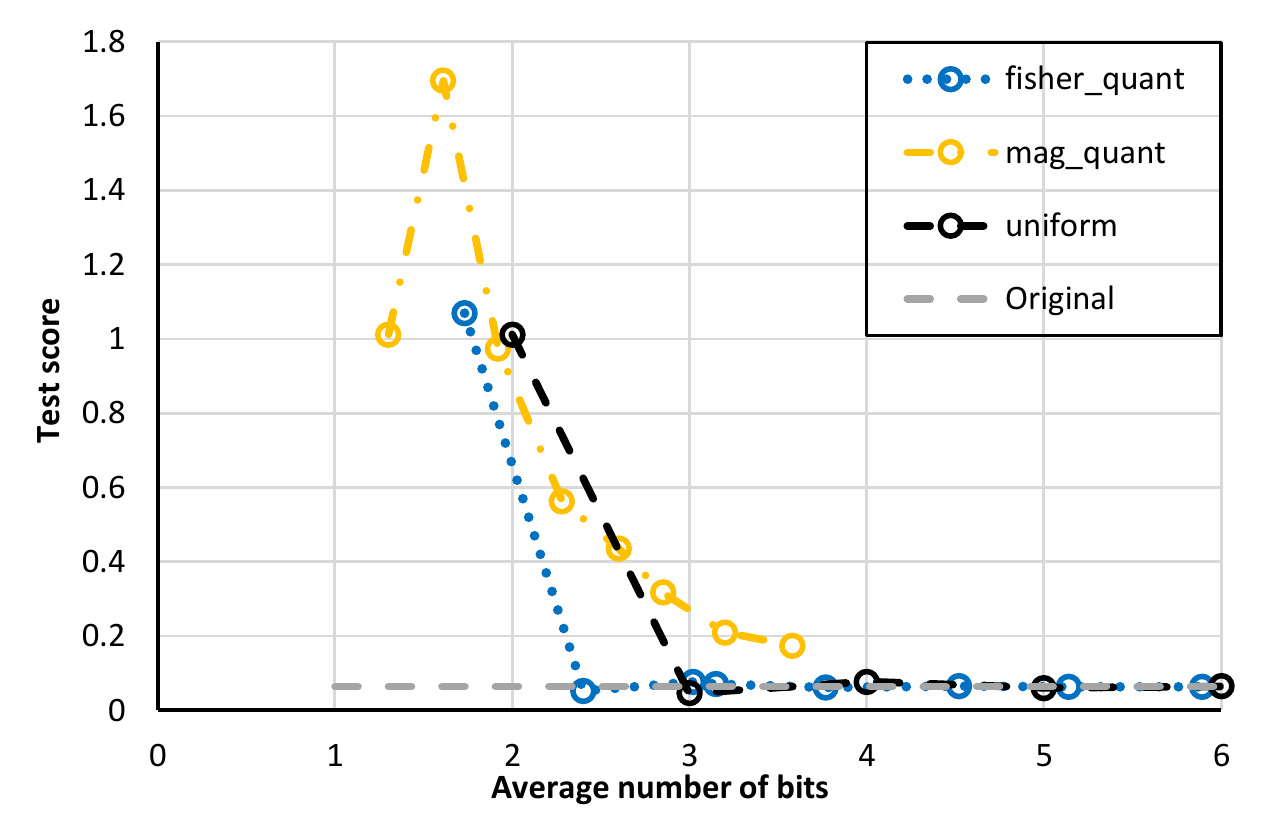}
  \caption{Scatter plot with line to show pairs of test score and average number of bits of different quantization methods.}\label{fig5}
\endminipage\hfill
\end{figure}

After pruning, the remaining parameters in the network must be quantized for fixed-point implementations. Our quantization method is a non-uniform quantization method based on $k$-means clustering. We rank-order the weights by an importance metric (``fisher" or ``mag") and cluster them into $k$ clusters. The clusters are then quantized using varying bit depths, from 1 bit/parameter (least important) to $k$ bits (most important).

To remove the effects of different pruning methods, we first removed $5.4\mathrm{E}5$ parameters (91.33\% of the total parameters) using magnitude-based pruning only. The resulting model had 51,242 (8.67\%) parameters remaining and an overall accuracy of 98.67\% (less than 1\% accuracy loss). We quantized the remaining parameters using three different methods: (1) non-uniform quantization based on Fisher Information ranking; (2) non-uniform quantization based on magnitude-based ranking; and (3) uniform quantization.  For methods (1) and (2),  we varied the number of clusters (from 3 to 10) and estimated the accuracy of the model for each value of $k$.

The results are shown in Fig. \ref{fig4} and Fig. \ref{fig5}: ``fisher\_quant'' represents non-uniform quantization based on Fisher Information, ``mag\_quant'' represents non-uniform quantization based on magnitude ranking, ``uniform'' represents uniform quantization and ``Original'' represents the original non-quantized model after removing $5.4\mathrm{E}5$ parameters (accuracy is 98.67\%).

%Then we assigned different number of bits to different clusters based on their importance such that clusters that were more important would be assigned more bits. For example, we had $K$ clusters, and we assigned 1 bit to the least important cluster and $K$ bits to the most important cluster. The third method, uniform quantization, just assigned the same number of bits to all remaining parameters. Note that the magnitude values of our model parameters are between 0 and 1, so besides 1 bit for sign the remaining bits are all used to quantize fractional part.

From the result, we see that non-uniform quantization based on the Fisher Information can achieve almost the same accuracy as the original model with only $2.4$ bits/parameter. This number is 3 for uniform quantization. Non-uniform quantization based on the magnitude never achieves the accuracy of the original model. The same pattern is seen for the test score. This shows that on this classification task, non-uniform quantization based on the Fisher Information achieves the highest compression ratio compared to non-uniform quantization based on magnitude ranking and uniform quantization.

\section{Discussion}
\label{discussion}

To evaluate the compression ratio for the example shown here, we analyze the effects of both network reduction and quantization. As we previously saw in Fig. \ref{fig2}, if we limit our acceptable reduction in performance to 1\%, then we can remove 92.18\% parameters (accuracy is 98.37\%) for magnitude based pruning. This results in a compression ratio of 12.8$\times$. For our proposed combination of magnitude and Fisher Information based pruning, we can remove 94.72\% parameters (accuracy is 98.38\%) and the compression ratio is 18.9$\times$.

For network quantization, we can choose between uniform or non-uniform quantization. If we assume the original parameters are saved in FLOAT32 format, as shown in Fig. \ref{fig4} using uniform quantization, we can achieve a reduction of $\frac{32}{3} = 10.7 \times$; by using non-uniform quantization, we can achieve a reduction of $\frac{32}{2.4} = 13.3 \times$. This means that the total compression ratio can be as high as 18.9 $\times$ 13.3 $\approx$ 251.4 with less than 1\% accuracy loss. This is likely an overestimate of the overall compression ratio because we need additional space to store the indices of the parameters that have been removed.

There is a relationship between our method and other previous methods based on estimation of the Hessian diagonal, namely optimal brain damage \cite{lecun1990optimal}, optimal brain surgeon \cite{hassibi1993second} and our previous work \cite{tu2016ranking}. The first two methods use the entries in the Hessian diagonal of the resulting cost function to identify important and unimportant parameters. These approaches are closely related to our approach when the cost function is the log-likelihood since the second derivative of log-likelihood function (Hessian) evaluated at the maximum likelihood estimate is the observed Fisher Information \cite{efron1978assessing}. Our approach in \cite{tu2016ranking} made use of the relationship between Fisher Information and the family of $f$-divergences to estimate the FIM diagonal. The principal difference between those approaches and the one we use here is scalability - the stochastic optimization method we use to estimate the FIM diagonal can be scaled to much larger network sizes.

Finally, it is important to note that further gains in performance can be obtained by retraining the network after pruning and before quantization \cite{han2015learning}. In this study, in an attempt to isolate the effects of network reduction and quantization, we elected not to retrain the network after the fact.

\section{Conclusion}
\label{conclusion}

In this paper, we propose a new network reduction and quantization scheme that uses a combination of the parameter magnitude and the Fisher Information as a measure of parameter importance. For network reduction, the proposed algorithm first removes parameters with small magnitude and then further reduces the network by removing additional parameters based on the Fisher Information. Following network reduction, we propose a non-uniform quantization scheme for remaining parameters based on the same Fisher criterion. The results show that the combination of network reduction and quantization results in large compression ratios. In future, our aim is to embed complexity reduction criteria in the training process instead of using it as a post-processing step.


\section*{Acknowledgment}

This research was supported in part by the Office of Naval Research grant N000141410722 (Berisha), an ASU-Mayo seed grant, and a hardware grant from NVIDIA.



%\bibliographystyle{IEEEtran}

\begin{thebibliography}{10}

\bibitem{Simonyan14c}
K.~Simonyan and A.~Zisserman,
\newblock ``Very deep convolutional networks for large-scale image
  recognition,''
\newblock {\em CoRR}, vol. abs/1409.1556, 2014.

\bibitem{szegedy2015going}
Christian Szegedy, Wei Liu, Yangqing Jia, Pierre Sermanet, Scott Reed, Dragomir
  Anguelov, Dumitru Erhan, Vincent Vanhoucke, and Andrew Rabinovich,
\newblock ``Going deeper with convolutions,''
\newblock in {\em Proceedings of the IEEE Conference on Computer Vision and
  Pattern Recognition}, 2015, pp. 1--9.

\bibitem{hinton2012deep}
Geoffrey Hinton, Li~Deng, Dong Yu, George~E Dahl, Abdel-rahman Mohamed, Navdeep
  Jaitly, Andrew Senior, Vincent Vanhoucke, Patrick Nguyen, Tara~N Sainath,
  et~al.,
\newblock ``Deep neural networks for acoustic modeling in speech recognition:
  The shared views of four research groups,''
\newblock {\em Signal Processing Magazine, IEEE}, vol. 29, no. 6, pp. 82--97,
  2012.

\bibitem{graves2014towards}
Alex Graves and Navdeep Jaitly,
\newblock ``Towards end-to-end speech recognition with recurrent neural
  networks,''
\newblock in {\em Proceedings of the 31st International Conference on Machine
  Learning (ICML-14)}, 2014, pp. 1764--1772.

\bibitem{lecun2015deep}
Yann LeCun, Yoshua Bengio, and Geoffrey Hinton,
\newblock ``Deep learning,''
\newblock {\em Nature}, vol. 521, no. 7553, pp. 436--444, 2015.

\bibitem{goldberg2015primer}
Yoav Goldberg,
\newblock ``A primer on neural network models for natural language
  processing,''
\newblock {\em arXiv preprint arXiv:1510.00726}, 2015.

\bibitem{gokhale2014240}
Vinayak Gokhale, Jonghoon Jin, Aysegul Dundar, Berin Martini, and Eugenio
  Culurciello,
\newblock ``A 240 g-ops/s mobile coprocessor for deep neural networks,''
\newblock in {\em Proceedings of the IEEE Conference on Computer Vision and
  Pattern Recognition Workshops}, 2014, pp. 682--687.

\bibitem{lecun1998gradient}
Yann LeCun, L{\'e}on Bottou, Yoshua Bengio, and Patrick Haffner,
\newblock ``Gradient-based learning applied to document recognition,''
\newblock {\em Proceedings of the IEEE}, vol. 86, no. 11, pp. 2278--2324, 1998.

\bibitem{krizhevsky2012imagenet}
Alex Krizhevsky, Ilya Sutskever, and Geoffrey~E Hinton,
\newblock ``Imagenet classification with deep convolutional neural networks,''
\newblock in {\em Advances in neural information processing systems}, 2012, pp.
  1097--1105.

\bibitem{cheng2015exploration}
Yu~Cheng, Felix~X Yu, Rogerio~S Feris, Sanjiv Kumar, Alok Choudhary, and Shi-Fu
  Chang,
\newblock ``An exploration of parameter redundancy in deep networks with
  circulant projections,''
\newblock in {\em Proceedings of the IEEE International Conference on Computer
  Vision}, 2015, pp. 2857--2865.

\bibitem{denil2013predicting}
Misha Denil, Babak Shakibi, Laurent Dinh, Nando de~Freitas, et~al.,
\newblock ``Predicting parameters in deep learning,''
\newblock in {\em Advances in Neural Information Processing Systems}, 2013, pp.
  2148--2156.

\bibitem{xue2013restructuring}
Jian Xue, Jinyu Li, and Yifan Gong,
\newblock ``Restructuring of deep neural network acoustic models with singular
  value decomposition.,''
\newblock in {\em INTERSPEECH}, 2013, pp. 2365--2369.

\bibitem{denton2014exploiting}
Emily~L Denton, Wojciech Zaremba, Joan Bruna, Yann LeCun, and Rob Fergus,
\newblock ``Exploiting linear structure within convolutional networks for
  efficient evaluation,''
\newblock in {\em Advances in Neural Information Processing Systems}, 2014, pp.
  1269--1277.

\bibitem{han2015learning}
Song Han, Jeff Pool, John Tran, and William Dally,
\newblock ``Learning both weights and connections for efficient neural
  network,''
\newblock in {\em Advances in Neural Information Processing Systems}, 2015, pp.
  1135--1143.

\bibitem{lecun1990optimal}
Yann LeCun, John~S Denker, and Sara~A Solla,
\newblock ``Optimal brain damage,''
\newblock in {\em Advances in Neural Information Processing Systems}, 1990, pp.
  598--605.

\bibitem{hassibi1993second}
Babak Hassibi and David~G Stork,
\newblock ``Second order derivatives for network pruning: Optimal brain
  surgeon,''
\newblock in {\em Advances in Neural Information Processing Systems}, 1993, pp.
  164--171.

\bibitem{yu2012exploiting}
Dong Yu, Frank Seide, Gang Li, and Li~Deng,
\newblock ``Exploiting sparseness in deep neural networks for large vocabulary
  speech recognition,''
\newblock in {\em Acoustics, Speech and Signal Processing (ICASSP), 2012 IEEE
  International Conference on}. IEEE, 2012, pp. 4409--4412.

\bibitem{han2015deep}
Song Han, Huizi Mao, and William~J Dally,
\newblock ``Deep compression: Compressing deep neural networks with pruning,
  trained quantization and huffman coding,''
\newblock {\em arXiv preprint arXiv:1510.00149}, 2015.

\bibitem{koksal2001weight}
Fatih K{\"o}ksal, Ethem Alpaydyn, and G{\"u}nhan D{\"u}ndar,
\newblock ``Weight quantization for multi-layer perceptrons using soft weight
  sharing,''
\newblock in {\em Artificial Neural Networks (ICANN) 2001}, pp. 211--216.
  Springer, 2001.

\bibitem{takeda2014boundary}
Ryu Takeda, Naoyuki Kanda, and Nobuo Nukaga,
\newblock ``Boundary contraction training for acoustic models based on discrete
  deep neural networks,''
\newblock in {\em INTERSPEECH}, 2014.

\bibitem{lei2013accurate}
Xin Lei, Andrew Senior, Alexander Gruenstein, and Jeffrey Sorensen,
\newblock ``Accurate and compact large vocabulary speech recognition on mobile
  devices.,''
\newblock in {\em INTERSPEECH}, 2013, pp. 662--665.

\bibitem{vanhoucke2011improving}
Vincent Vanhoucke, Andrew Senior, and Mark~Z Mao,
\newblock ``Improving the speed of neural networks on cpus,''
\newblock in {\em Proc. Deep Learning and Unsupervised Feature Learning NIPS
  Workshop}, 2011, vol.~1.

\bibitem{gupta2015deep}
Suyog Gupta, Ankur Agrawal, Kailash Gopalakrishnan, and Pritish Narayanan,
\newblock ``Deep learning with limited numerical precision,''
\newblock {\em arXiv preprint arXiv:1502.02551}, 2015.

\bibitem{gong2014compressing}
Yunchao Gong, Liu Liu, Ming Yang, and Lubomir Bourdev,
\newblock ``Compressing deep convolutional networks using vector
  quantization,''
\newblock {\em arXiv preprint arXiv:1412.6115}, 2014.

\bibitem{soulie2015compression}
Guillaume Souli{\'e}, Vincent Gripon, and Ma{\"e}lys Robert,
\newblock ``Compression of deep neural networks on the fly,''
\newblock {\em arXiv preprint arXiv:1509.08745}, 2015.

\bibitem{wang2015small}
Yongqiang Wang, Jinyu Li, and Yifan Gong,
\newblock ``Small-footprint high-performance deep neural network-based speech
  recognition using split-vq,''
\newblock in {\em Acoustics, Speech and Signal Processing (ICASSP), 2015 IEEE
  International Conference on}. IEEE, 2015, pp. 4984--4988.

\bibitem{tu2016ranking}
Ming Tu, Visar Berisha, Martin Woolf, Jae-sun Seo, and Yu~Cao,
\newblock ``Ranking the parameters of deep neural network using the fisher
  information,''
\newblock in {\em Acoustics, Speech and Signal Processing (ICASSP), 2016 IEEE
  International Conference on}. IEEE, under publication, 2016.

\bibitem{amari1998natural}
Shun-Ichi Amari,
\newblock ``Natural gradient works efficiently in learning,''
\newblock {\em Neural computation}, vol. 10, no. 2, pp. 251--276, 1998.

\bibitem{pascanu2013revisiting}
Razvan Pascanu and Yoshua Bengio,
\newblock ``Revisiting natural gradient for deep networks,''
\newblock {\em arXiv preprint arXiv:1301.3584}, 2013.

\bibitem{povey2014parallel}
Daniel Povey, Xiaohui Zhang, and Sanjeev Khudanpur,
\newblock ``Parallel training of dnns with natural gradient and parameter
  averaging,''
\newblock {\em arXiv preprint arXiv:1410.7455}, 2014.

\bibitem{desjardins2015natural}
Guillaume Desjardins, Karen Simonyan, Razvan Pascanu, et~al.,
\newblock ``Natural neural networks,''
\newblock in {\em Advances in Neural Information Processing Systems}, 2015, pp.
  2062--2070.

\bibitem{kingma2014adam}
Diederik Kingma and Jimmy Ba,
\newblock ``Adam: A method for stochastic optimization,''
\newblock {\em arXiv preprint arXiv:1412.6980}, 2014.

\bibitem{chollet2015keras}
François Chollet,
\newblock ``Keras,'' https://github.com/fchollet/keras, 2015.

\bibitem{zeiler2014visualizing}
Matthew~D Zeiler and Rob Fergus,
\newblock ``Visualizing and understanding convolutional networks,''
\newblock in {\em Computer vision--ECCV 2014}, pp. 818--833. Springer, 2014.

\bibitem{efron1978assessing}
Bradley Efron and David~V Hinkley,
\newblock ``Assessing the accuracy of the maximum likelihood estimator:
  Observed versus expected fisher information,''
\newblock {\em Biometrika}, vol. 65, no. 3, pp. 457--483, 1978.

\end{thebibliography}

\end{document}